\documentclass{article}

\usepackage{microtype}
\usepackage{graphicx}
\usepackage{subfigure}
\usepackage{booktabs} 

\usepackage{hyperref}

\usepackage[accepted]{icml2024}

\usepackage{amsmath}
\usepackage{amssymb}
\usepackage{mathtools}
\usepackage{amsthm}
\usepackage{multirow}
\usepackage{svg}
\usepackage{makecell}

\usepackage[capitalize,noabbrev]{cleveref}

\theoremstyle{plain}

\theoremstyle{definition}

\theoremstyle{remark}

\newcommand{\bftab}{\fontseries{b}\selectfont}
\newcommand{\und}{\underline}

\newcommand{\mname}{\textsc{AutoDiff}}

\usepackage[textsize=tiny]{todonotes}

\icmltitlerunning{\mname: Autoregressive Diffusion Modeling for Structure-based Drug Design}

\begin{document}

\twocolumn[
\icmltitle{\mname: Autoregressive Diffusion Modeling for Structure-based Drug Design}



\icmlsetsymbol{equal}{*}

\begin{icmlauthorlist}
\icmlauthor{Xinze Li}{galixir,equal}
\icmlauthor{Penglei Wang}{galixir}
\icmlauthor{Tianfan Fu}{rpi}
\icmlauthor{Wenhao Gao}{mit}
\icmlauthor{Chengtao Li}{galixir}
\icmlauthor{Leilei Shi}{galixir}
\icmlauthor{Junhong Liu}{galixir,equal}
\end{icmlauthorlist}

\icmlaffiliation{galixir}{Galixir Technologies}
\icmlaffiliation{rpi}{Department of Computer Science, Rensselaer Polytechnic Institute}
\icmlaffiliation{mit}{Department of Chemical Engineering, Massachusetts Institute of Technology}

\icmlcorrespondingauthor{Junhong Liu}{junhong.liu@galixir.com}

\icmlkeywords{Machine Learning, ICML}

\vskip 0.3in
]



\printAffiliationsAndNotice{\icmlEqualContribution} 

\begin{abstract}
Structure-based drug design (SBDD), which aims to generate molecules that can bind tightly to the target protein, is an essential problem in drug discovery, and previous approaches have achieved initial success. However, most existing methods still suffer from invalid local structure or unrealistic conformation issues, which are mainly due to the poor leaning of bond angles or torsional angles. To alleviate these problems, we propose \mname, a diffusion-based fragment-wise autoregressive generation model. Specifically, we design a novel molecule assembly strategy named conformal motif that preserves the conformation of local structures of molecules first, then we encode the interaction of the protein-ligand complex with an $SE(3)$-equivariant convolutional network and generate molecules motif-by-motif with diffusion modeling. In addition, we also improve the evaluation framework of SBDD by constraining the molecular weights of the generated molecules in the same range, together with some new metrics, which make the evaluation more fair and practical. Extensive experiments on CrossDocked2020 demonstrate that our approach outperforms the existing models in generating realistic molecules with valid structures and conformations while maintaining high binding affinity.

\end{abstract}

\section{Introduction}
\label{introduction}
Structure-based drug design (SBDD), which can be formulated as generating 3D molecules conditioned on protein pockets, is an important and challenging task in drug discovery \citep{bohacek1996art}. Compared to string-based~\cite{Bjerrum2017_MolRNN, Segler2018_GenRNN} and graph-based~\cite{You2018_GraphRNN, Jin2018_JTVAE, Shi2020_GraphAF, Jin2020_HTJVAE} molecule generation, SBDD leverages the spatial geometric structure information and perceives how molecule interacts with protein pocket. Therefore, it can generate drug-like molecules with high binding affinities to the target. Recently, we have witnessed the success of deep generative models on this task, and most of the existing approaches can be roughly divided into two categories: autoregressive-based and diffusion-based.

For autoregressive-based approaches, early attempts generated 3D molecules by estimating the probability density of atoms' occurrence in protein pocket and placing atoms of specific types and locations one by one~\cite{Luo2021_AR, Liu2022_GraphBP}. Subsequently, \citet{Peng2022_Pocket2Mol} took the modeling of chemical bonds into consideration and achieved more practical atomic connections. However, atom-wise autoregressive approaches always force the model to generate chemically invalid intermediaries~\cite{Jin2018_JTVAE}, yielding unrealistic fragments in the generated molecules. To tackle this problem, fragment-wise autoregressive approaches~\cite{Zhang2022_FLAG, Zhang2023_DrugGPS} were proposed, while these methods always suffer from error accumulation due to the poor learning capacity of torsional angle and the defective motif design strategy, which lead to invalid local structures or unrealistic conformations.

For diffusion-based approaches, some learned the distribution of atom types and positions from a standard Gaussian prior based on diffusion process~\cite{Guan2023_TargetDiff, Lin2022_Diffbp, Schneuing2022_DiffSBDD}, and some introduced the scaffold-arm decomposition prior into the diffusion modeling to improve the binding affinity of the generated molecules~\cite{Guan2023_DecompDiff}. However, diffusion-based approaches also tend to generate unrealistic local structures such as messy rings~\cite{Guan2023_DecompDiff}. In addition to SBDD, diffusion models have also been widely used in other biochemistry tasks such as molecule conformation prediction~\cite{Jing2022_TorsionalDiff} and ligand-protein binding prediction~\cite{Corso2022_DiffDock, Lu2023_DynamicBind}, where diffusion models show promising modeling capacity of torsional angle.

To overcome the aforementioned challenges and limitations, we leverage the strength of diffusion models and motif-based autoregressive generation and propose \mname, a novel conformal motif-based molecule generation method with diffusion modeling. Different from previous approaches~\cite{Zhang2022_FLAG}, we propose a novel conformal motif design strategy, which can alleviate the invalid structure and unrealistic conformation problems. In addition, we model the protein-ligand complex with an $SE(3)$-equivariant convolutional network to learn the spatial geometric structure features and interaction information. At each generation step, we predict a connection site which can be either an atom or a bond for the current fragment and the motif library, respectively, then attach two predicted connection sites to form a new fragment, and the torsional angle is predicted with a probabilistic diffusion model at last. Thanks to the implicitly encoded conformation in the conformal motifs, the connection site-based attachment can perceive the local environment of the current pocket-ligand complex, therefore alleviating the error accumulation and generating more realistic molecules. Furthermore, we also improve the evaluation framework by constraining the molecular weights of the generated molecules in the same range, together with some new metrics, which can evaluate the structure validity and binding affinity more practically than before.
 
To summarize, the main contributions of this paper are three-fold:
 \begin{itemize}
     \item \textbf{Assembly strategy}: we propose a new motif design strategy named conformal motif, which preserves all conformation information of local structures.
     \item \textbf{Generative method}: we present a novel generation framework which makes use of the advantages of diffusion model and motif-based generation to design realistic molecules.
     \item \textbf{Experimental result}: we improve the evaluation framework together with some new metrics, with which the SBDD models can be evaluated and compared more fairly and practically.
 \end{itemize}

\section{Related Work}
\label{related work}
\noindent\textbf{Fragment-Wise Molecule Generation.} Fragment-wise generation is prevalent since the chemical information is preserved in the substructures to produce realistic molecules. \citet{Jin2018_JTVAE} proposed a junction tree variational autoencoder for generating molecules with chemical motifs, it constructs a tree-structured scaffold first, and then combines the motifs of the tree into a molecule with a graph message passing network. \citet{Jin2020_ReationaleRL} designed a multiple-property optimization approach in which the motif vocabulary with good properties is constructed first, then molecules are generated by expanding rationale graphs with graph generative models and optimized by fine-tuning to desirable properties with reinforcement learning models. Recently, fragment-based 3D generation approaches have shown promising capacity in drug design and lead optimization~\cite{Flam2022_SFRL, Powers2022_PLS, Zhang2022_FLAG, Zhang2023_DrugGPS}. \citet{Flam2022_SFRL} used a hierarchical agent to generate 3D molecules guided by quantum mechanics with a reinforcement learning framework in an autoregressive fashion. \citet{Powers2022_PLS} learned how to attach fragments to a growing structure by recognizing realistic intermediates generated \textit{en route} to a final ligand, which solved a 3D molecule optimization problem. For fragment-wise generation, the key is how to design the motif that can encode the chemical information and local topological structure appropriately. This is even more important to structure-based drug design, which needs to take the motif conformation into account to achieve realistic 3D structures.

\noindent\textbf{Structure-Based Drug Design}. Structure-based drug design (SBDD) generates target-aware molecules that bind to specific protein pockets. \citet{fu2022reinforced} proposed a variant of genetic algorithm guided by reinforcement learning, which employs neural models to prioritize the profitable drug design steps with protein structure information as input. \citet{Ragoza2022_LiGAN} presented an atomic density grid representation of protein-ligand complex and learned the molecule distributions with a conditional variational autoencoder. \citet{Luo2021_AR, Liu2022_GraphBP} generated 3D molecules by estimating the probability density of atoms' occurrence in protein pocket and placing atoms of specific types and locations one by one. \citet{Peng2022_Pocket2Mol} took the modeling of chemical bonds into consideration and achieved more practical atomic connections.~\citet{Zhang2022_FLAG, Zhang2023_DrugGPS} proposed a fragment-wise framework which generates molecules motif-by-motif. Another line of work focuses on diffusion-based approaches.~\citet{Lin2022_Diffbp, Schneuing2022_DiffSBDD, Guan2023_TargetDiff} learned the distribution of atom types and positions from a standard Gaussian prior based on the diffusion process. \citet{Guan2023_DecompDiff} decomposed ligands into arms and scaffolds, then incorporated related prior knowledge into diffusion models for better molecule generation. 

\noindent\textbf{Diffusion Models}. Recently, diffusion models have attracted considerable attention thanks to their promising generative results, which have been widely used in computer vision~\cite{Dhariwal2021_beatGAN, Nichol2021_Glide, Rombach2022_StableDiff, Ceylan2023_Pix2video, Tumanyan2023_PPD}, natural language processing~\cite{Li2022_DiffLM, Lovelace2022_Ld4LG, Yuan2022_Seqdiffuseq, Lin2023_GENIE}, and speech modeling~\cite{Pascual2023_DAG, Guo2023_Emodiff}, while remarkable success also has been achieved in the domain of biochemistry and drug design~\cite{Hoogeboom2022_EDM, Xu2022_GeoDiff, Jing2022_TorsionalDiff, Corso2022_DiffDock, Lu2023_DynamicBind}. \citet{Jing2022_TorsionalDiff} studied molecular conformation generation, which operates on the space of torsional angles via a diffusion process on the hypertorus and an extrinsic-to-intrinsic score model. \citet{Hoogeboom2022_EDM} generated 3D molecules, which learns to denoise a diffusion process with an equivariant network that jointly operates on both atom coordinates and atom types.

\section{Method}
\label{method}
In this section, we present \mname, a diffusion-based fragment-wise autoregressive generation model for structure-based drug design. Firstly, we formulate the task of structure-based drug design (SBDD) formally in \cref{probelm_formulation} and introduce conformal motif design strategy in \cref{sec:conformal_motif}. Then, we elaborate on the generation process based on the proposed conformal motif in \cref{tiffmlp}. In the end, we derive the optimization objective to train our model in \cref{training}.

\begin{algorithm}[bt]
\caption{Conformal Motif Extraction (Section~\ref{sec:conformal_motif})}
\label{motif_alg}
\begin{algorithmic}
   \STATE {\bfseries Input:} A set of molecule graphs $\mathcal{D}=\{\mathcal{G}_1, \mathcal{G}_2,...,\mathcal{G}_{|\mathcal{D}|}\}$ 
   \STATE {\bfseries Output:} Motif vocabulary $\mathcal{W}$
   \STATE $\mathcal{W}_r \gets \{\}, \mathcal{W}_{c*} \gets \{\}$
   \FOR{$\mathcal{G}(\mathcal{V}, \mathcal{E}) \in \mathcal{D}$}
   \STATE $\mathcal{W}_{\mathcal{G}}$ = Disconnect($\mathcal{G}, \mathcal{R}_{\mathcal{G}})$
       \FOR{$\mathcal{F}(\mathcal{V}_{\mathcal{F}}, \mathcal{E}_{\mathcal{F}})  \in \mathcal{W}_{\mathcal{G}}$}
            \IF{IsFusedRing($\mathcal{F})$}
            \STATE $\mathcal{F}_{r}$ = Decompose($\mathcal{F}$)
            \STATE $\mathcal{W}_r \gets \mathcal{W}_r \cup \mathcal{F}_{r}$
            \ELSIF{IsChain($\mathcal{F})$}
            \STATE $\mathcal{V}_{\mathcal{F}}^{*} \gets \mathcal{V}_{\mathcal{F}} \cup \{a|a\in \mathcal{V}, \exists b\in \mathcal{V}_{\mathcal{F}}, (a, b) \in \mathcal{E} \}$
            \STATE $\mathcal{E}_{\mathcal{F}}^{*} \gets \{(a, b) \in \mathcal{E} | a\in\mathcal{V}_{\mathcal{F}}^{*}, b \in \mathcal{V}_{\mathcal{F}}^{*} \}$
            \STATE $\mathcal{W}_{c*} \gets \mathcal{W}_{c*} \cup \{ \mathcal{F}^{*}(\mathcal{V}_{\mathcal{F}}^{*}, \mathcal{E}_{\mathcal{F}}^{*}) \}$
            \ELSE
            \STATE $\mathcal{W}_r \gets \mathcal{W}_r \cup \{\mathcal{F}\}$
            \ENDIF  
       \ENDFOR
   \ENDFOR
   \STATE $\mathcal{W} \gets \mathcal{W}_r \cup \mathcal{W}_{c*}$
\end{algorithmic}
\end{algorithm}

\subsection{Problem Formulation}
\label{probelm_formulation}

Structure-based drug design (SBDD) task can be formulated as a conditional generation task that generates 3D molecules conditioned on the given protein pocket. Specifically, the protein pocket can be represented as a set of atoms (with coordinates)
$\mathcal{P} =\{(a_P^{i}, \mathbf{r}_P^{i})\}_{i=1}^{N_P}$, while the drug molecule can also be represented as a set of atoms $\mathcal{G} =\{(a_G^{i}, \mathbf{r}_G^{i})\}_{i=1}^{N_{G}}$, where $N_P$ and 
$N_G$ denotes the number of atoms in the pocket and the molecule, respectively; 
$\mathbf{r}^{i} \in \mathbb{R}^3$ is the coordinate of the $i$-th heavy atom. With the definitions, the SBDD task can be re-formulated as learning a conditional distribution $p(\mathcal{G}|\mathcal{P})$ from the co-crystallized (or docked) 3D protein-ligand complex data.

\subsection{Assembly: Conformal Motif}
\label{sec:conformal_motif}
\begin{figure}[t]
\vskip 0.2in
\begin{center}
\centerline{\includegraphics[width=\columnwidth]{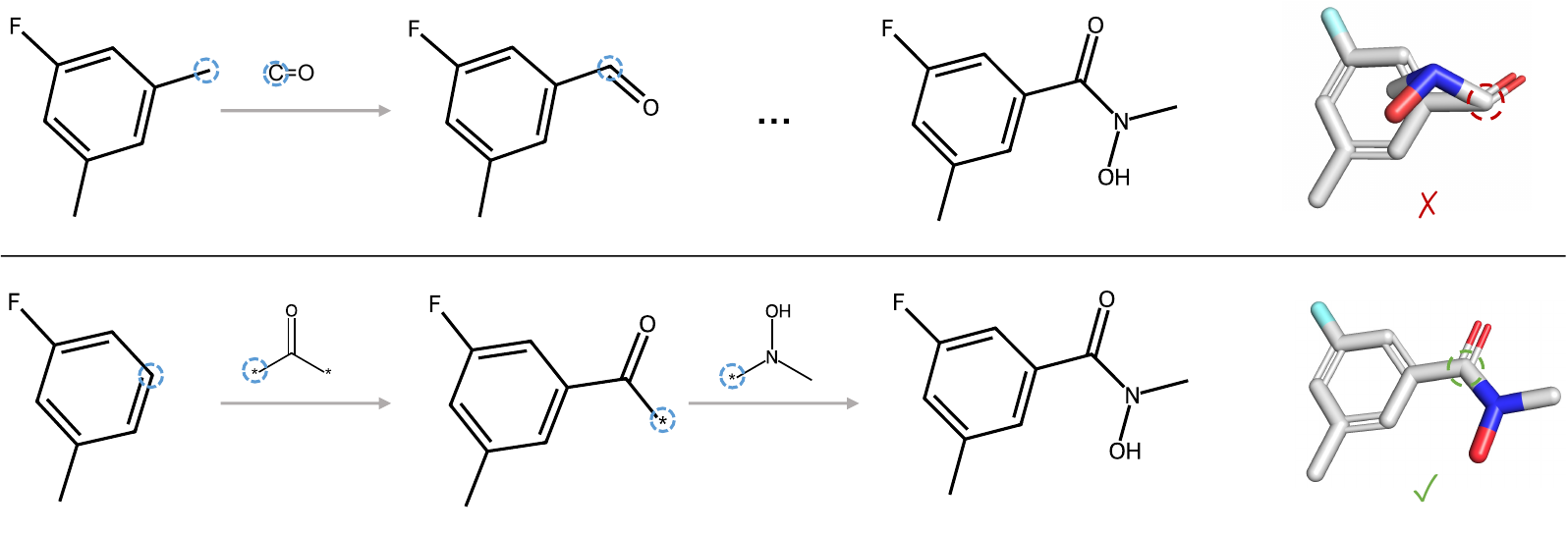}}
\caption{Illustration of the advantage of conformal motif (bottom) versus other methods~\cite{Zhang2022_FLAG, Zhang2023_DrugGPS} (top).}
\label{motifcmp-png}
\end{center}
\vskip -0.2in
\end{figure}

The motif-based generation is explored and applied in 2D generation~\cite{Jin2018_JTVAE, Jin2020_HTJVAE, Jin2020_ReationaleRL, fu2021mimosa} initially, and has achieved decent performance especially compared to atom-wise approaches. 
To construct motif vocabulary, molecules in a library are decomposed into disconnected fragments by breaking all the bridge bonds or rotatable bonds that will not violate chemical validity, and fragments with higher frequency than a threshold are selected as the building blocks, i.e., motifs. This strategy was also employed in 3D generation~\cite{Flam2022_SFRL, Powers2022_PLS} and SBDD~\cite{Zhang2022_FLAG, Zhang2023_DrugGPS}, while the results are not satisfied due to the invalid structures and conformations generated in the sampled molecules. The main reason is that the existing motif design strategy is defective since it only encodes part of the 3D topological information of local structures. Specifically, some 3D topological information of the surrounding environment of atoms in severed bonds will be lost during fragmentation, which leads to the annihilation of the correct local conformation when motifs are attached to each other and finally results in invalid structures or unrealistic conformations, as shown in \cref{motifcmp-png}. Motivated by the analysis results, we propose a novel motif design strategy, i.e., conformal motif, in which the term ``conformal'' stems from hydromechanics and geometry, while we refer to fully preserving 3D topology information in our motifs. Concretely, we first detach all freely rotatable bonds $\mathcal{R}$ (precise definitions in \cref{appendix:rotatable_bond}) to break molecules into fragments, then we use redundant dummy atoms to act as placeholders,  which preserve the 3D topology information (mainly bond angles) of the surrounding environment implicitly for atoms of the severed bonds, and conformation of the motifs can be recovered with cheminformatics tools such as RDKit~\cite{Bento2020_RDKit}. \cref{motifcmp-png} shows that it avoids distorting local structures and helps to generate molecules with realistic structures and conformations. Furthermore, to explore more possible conformation flexibly, we further decompose the fused ring to reduce the motif size. The extracted conformal motifs can be divided into two categories: ring-like $\mathcal{W}_{r}$ and chain-like $\mathcal{W}_{c*}$. In view of the combination explosion of adding dummy atoms on $\mathcal{W}_{r}$, we only add dummy atoms on $\mathcal{W}_{c*}$. \cref{motif_alg} shows the pseudo-code of the complete process. We also provide an example in Appendix~\ref{appendix:motif_example}.  To the best of our knowledge, conformal motif is the first motif strategy designed for the SBDD task that takes full conformation information into consideration.

\begin{figure*}[ht]
\vskip 0.2in
\begin{center}
\centerline{\includegraphics[width=.95\textwidth]{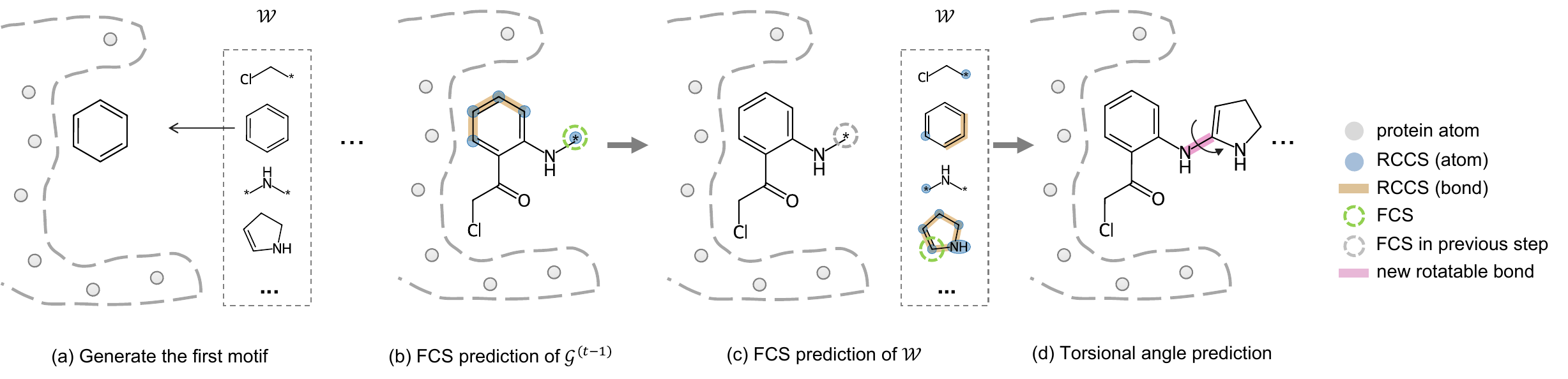}}
\caption{Overview of the generation process of \mname. RCCS: Reduced Candidate Connection Site. FCS: Focal Connection Site. Details are shown in \cref{tiffmlp}.}
\label{tiffmlp-png}
\end{center}
\vskip -0.2in
\end{figure*}

\subsection{Method: \mname}
\label{tiffmlp}
We first present the overall generation process of the proposed \mname, together with some notion definitions in \cref{sec:overview}; then we show the $SE(3)$-equivariant encoder in \cref{sec:encoder}, which is employed as the main architecture to model the protein pocket-ligand complex; finally, we introduce each module from \cref{sec:first_motif} to \cref{sec:torsional} in detail. For ease of exposition, we list all the terminologies and mathematical notations in~\cref{appendix:term}.

\subsubsection{Overview of \mname}
\label{sec:overview}
Firstly, we will define some notions. A connection site is an atom in a severed bond or a bond in a fused ring during motif extraction. A candidate connection site is an atom or a bond that may become a connection site during the generation process. We denote all the candidate connection sites in the intermediate $\mathcal{G}^{(t)}$ or the motif vocabulary $\mathcal{W}$ as CCS, while some atoms or bonds in CCS of $\mathcal{W}$ are chemically equivalent and can be reduced to a simplified version, which is called Reduced Candidate Connection Sites (RCCS). It is worth noting that RCCS is the same as CCS in $\mathcal{G}^{(t)}$. The item selected from RCCS for attachment in each generation step is termed as Focal Connection Site (FCS).

Overall,  the process of which can be defined as follows:
\begin{align}
& \mathcal{G}^{(t)} = \phi(\mathcal{P}), & t=1, \\ 
& \mathcal{G}^{(t)} = \phi(\mathcal{G}^{(t-1)}, \mathcal{P}), & t>1, 
\end{align}
where $\phi$ is our generation model. The generation of each motif consists of four steps (\cref{tiffmlp-png}): (1) an FCS prediction model is trained to predict the FCS in $\mathcal{G}^{(t-1)}$; (2) another FCS prediction model predicts the FCS in $\mathcal{W}$, and the corresponding motif $\mathcal{W}^{(t-1)}$ is also determined simultaneously; (3) $\mathcal{W}^{(t-1)}$ will be attached to $\mathcal{G}^{(t-1)}$ by connecting the two FCSs; (4) learn the torsional angle with a diffusion-based model. In this way, we get $\mathcal{G}^{(t)}$ and the generation process continues until no FCS in the current ligand fragment can be found. Note that the generation of the first motif is different from the procedure presented above and we will introduce the details in \cref{sec:first_motif}.

\subsubsection{Contextual Encoder}
\label{sec:encoder}
This section describes an $SE(3)$-equivariant convolutional network-based encoder, which is employed as the main architecture to model the protein pocket-ligand complex. 
It is crucial to characterize the surrounding environmental information in the protein pocket for the SBDD task. In our approach, complexes of protein pockets and ligand fragments are collectively represented as heterogeneous geometric graphs
$\mathcal{G}_H = (\mathcal{V}_H, \mathcal{E}_H)$, where the vertex set $\mathcal{V}_H = (\mathcal{V}_l, \mathcal{V}_p)$ is the collection of all the heavy atoms of ligand fragment and protein pocket, while the edge set $\mathcal{E}_H=(\mathcal{E}_{ll}, \mathcal{E}_{lp}, \mathcal{E}_{pl}, \mathcal{E}_{pp})$ is constructed by cutting off the distance between atoms with thresholds of 5$\mathring{A}$\footnote{The units is Angstrom $\mathring{A}$ ($10^{-10}$ m).}, 10$\mathring{A}$, 15$\mathring{A}$ for ligand-ligand, ligand-pocket/pocket-ligand, and pocket-pocket atom pairs respectively. Unlike previous approaches~\cite{Peng2022_Pocket2Mol, Zhang2022_FLAG} that build interaction graphs only depending on distance or $k$-nearest neighbors, we also preserve all the covalent bonds in ligands as edges to better model ligand-ligand atoms interactions.

$SE(3)$-equivariant convolutional networks based on tensor products of irreducible representations of $SO(3)$ are used to encode $\mathcal{G}_H$. At each interaction layer, messages are generated using a tensor product of spherical harmonic representations of the edge vector and node representation. Then, for each node $i$ of type $c_a (c_a\in\{l, p\})$, it collects the message from its connected edges and updates its representation, which can be formulated as: 
\begin{equation}
    \mathbf{h}_i \leftarrow \mathbf{h}_i \underset{c\in\{l,p\}}{\oplus} \mathrm{BN}^{(c_a, c)}\left(\frac{1}{|\mathcal{N}_i^{(c)}|} \sum_{j\in\mathcal{N}_i^{(c)}} Y(\mathbf{r}_{ij}) \otimes_{\psi_{ij}} \mathbf{h}_j \right), 
\end{equation}
where $\mathbf{h}_i$ denotes node $i$'s representation, $\oplus$ denotes vector addition, $\otimes_{\psi_{ij}}$ denotes spherical tensor product with weight $\psi_{ij}$, $\mathrm{BN}$ is the equivariant batch normalization, $\mathcal{N}_i^{(c)}$ denotes neighbors of node $i$ of type $c$. $Y$ refer to spherical harmonics, $\mathbf{r}_{ij}$ is the direction vector of edge $e_{ij}$, $\psi_{ij} = \Psi(\mathrm{h}_{ij}, \mathrm{h}_j, \mathrm{h}_j)$ contains learnable weight of tensor product, where $\mathrm{h}_{ij}$ denotes the embedding of $e_{ij}$, and $\mathrm{h}_j$ denotes scalar features of node $i$. Note that the interaction layer contains three sublayers: two intra-interaction layers on $\mathcal{G}=(\mathcal{V}_l, \mathcal{E}_{ll})$ and   $\mathcal{P}=(\mathcal{V}_p, \mathcal{E}_{pp})$, one inter-interaction layer on $((\mathcal{V}_l, \mathcal{V}_p), (\mathcal{E}_{pl}, \mathcal{E}_{lp}))$.

\subsubsection{Generate the First Motif}
\label{sec:first_motif}
How to generate the first motif is crucial to achieving successful generation, which consists of two steps: selection of the motif and placement in the protein pocket. We first train a model to predict the frontier in the pocket, which is defined as the pocket atom closest to the first motif’s centroid, and then a classifier is employed to predict the motif by taking the predicted frontier as input. So far, we have selected the first motif, while placing it in the pocket, namely pose prediction, is quite challenging. Previous methods used contact map to predict the position for the first motif~\cite{Zhang2022_FLAG, Zhang2023_DrugGPS}, which tends to predict implausible poses and mislead the whole generation process due to the inappropriate modeling of the motif poses. To be specific, the same motif may have different relative positions to the same pocket or subpocket in different complexes, while the contact map-based approaches are forced to learn an average-like output from the training data consisting of various alternative poses, thus usually resulting in unrealistic molecule structures. In our approach, we develop a
 diffusion-based generative model to learn the
 distribution of various motif poses. 
Since there are no freely rotatable bonds in the motif, the pose lies in a 6-dimensional submanifold whose degree of freedom comes from translation and rotation. Therefore, we generate the conformation of the selected motif by RDKit~\cite{Bento2020_RDKit} first,  and a convolution of each motif atom with motif centroid $o$ is employed:

\begin{equation}
     \mathbf{v} = \frac{1}{|\mathcal{V}_l|} \sum_{i\in \mathcal{V}_l} Y(\mathbf{r}_{oi})  \otimes_{\psi_{oi}} \mathbf{h}_i,
\end{equation}
where $\psi_{oi}= \Psi(\mathrm{h}_{oi}, \mathrm{h}_i)$. The output $\mathbf{v}$ consists of 2 odd parity vectors and 2 even vectors. Translation and rotation of the molecule are predicted as:
\begin{align}
    \mathbf{tr} &= \frac{\mathbf{\bar{v}}_{odd}}{\left\| \mathbf{\bar{v}}_{odd} \right\|} \times \mathrm{MLP} \left( \left\| \mathbf{\bar{v}}_{odd} \right\|, \mathbf{s}_t \right), \\
    \mathbf{rot} &= \frac{\mathbf{\bar{v}}_{even}}{\left\| \mathbf{\bar{v}}_{even} \right\|} \times \mathrm{MLP} \left( \left\| \mathbf{\bar{v}}_{even} \right\|, \mathbf{s}_t \right), 
\end{align}
where $\mathbf{s}_t$ denotes the sinusoidal embeddings of the diffusion time ${t}$.

\subsubsection{FCS Prediction}
\label{sec:fcs}
The key step for fragment-wise autoregressive generation is selecting a motif $\mathcal{W}^{(t-1)}$ and attaching it to the current generated molecule $\mathcal{G}^{(t-1)}$. Different from previous methods~\cite{Zhang2022_FLAG, Zhang2023_DrugGPS} that predict a motif first and then select the appropriate attachment by enumeration and scoring, we predict a connection site directly in the conformal motif vocabulary $\mathcal{W}$, while $\mathcal{W}^{(t-1)}$ is determined simultaneously. In this way, the motif is predicted in a more fine-grained fashion, taking the atom-level contextual information into account. In the following, we will elaborate on FCS prediction of $\mathcal{G}^{(t-1)}$ and $\mathcal{W}$, respectively.

\noindent\textbf{FCS prediction of $\mathcal{G}^{(t-1)}$}. According to the definition in \cref{sec:overview}, CCS for $\mathcal{G}^{(t-1)}$ includes three parts: dummy atoms in chains, atoms in rings that do not violate the chemical validity if connected to other heavy atoms, and chemical bonds in rings that connect two candidate connection atoms. Therefore, we train a model to predict the FCS from CCS as follows:
    \begin{align}
    P_{v_i} &= \sigma(\mathrm{MLP}(\mathbf{h}_i)), \\
    P_{e_u} &= \sigma(\mathrm{MLP}([\mathbf{h}_i, \mathbf{h}_j])), (v_i, v_j) \in e_u. 
    \end{align}
    Note that edges are directed, and the predicted FCS can be an atom $v_f$ or an edge $e_f$.
    
\noindent\textbf{FCS prediction of $\mathcal{W}$}. 
As defined in \cref{sec:overview}, the construction of CCS for $\mathcal{W}$ is similar to $\mathcal{G}^{(t-1)}$ which also consists of three parts: dummy atoms in $\mathcal{W}_{c*}$, atoms in $\mathcal{W}_{r}$ that do not violate the chemical validity if connected to other heavy atoms, and bonds in $\mathcal{W}_{r}$ that connect two candidate connection atoms. While some atoms or bonds in CCS are equivalent, which means that their centered neighbors of atoms and bonds are all the same, resulting in the same graph generated by connecting them to $\mathcal{G}^{(t-1)}$. Therefore, equivalent atoms or bonds should be reduced in case of inducing aleatoric uncertainty in generation. To reduce the CCS into the RCCS, we first recognize equivalent atoms and bonds. Specifically, we traverse all the atom pairs $(v_i, v_j)$ in a motif and define corresponding graph pairs $(\mathcal{G}_i^{\mathcal{W}}, \mathcal{G}_j^{\mathcal{W}})$, which denote the motif graph centered in $v_i$ and $v_j$, In the graph pair, $v_i$ and $v_j$ are assigned with a special label, while the other atoms and bonds are labeled with their corresponding element type or bond type, respectively. If the graph pairs $(\mathcal{G}_i^{\mathcal{W}}, \mathcal{G}_j^{\mathcal{W}})$ are proven to be isomorphic under the graph isomorphism testing, the atoms $v_i$ and $v_j$ are equivalent. On the basis of this, equivalent edges are determined by whether their connected atoms are equivalent, in other words, $ e_{ij} \equiv e_{mn} \iff (v_i \equiv v_m)  \land (v_j \equiv v_n)$, where $\equiv$ refers to equivalent. After recognizing the equivalent connection sites, we reduce the CCS into the RCCS. In the end, we use FCS in $\mathcal{G}^{(t-1)}$ to query another FCS in $\mathcal{W}$, and employ two neural networks to make a query vector $Q$ and key vectors $K$. Specifically, FCS is predicted by either of the following models:
\begin{align}
    P_v &= \underset{v\in \mathcal{V}_{\mathcal{W}}}{\mathrm{softmax}} \left(\mathrm{MLP}_Q^v ([\mathbf{h}_{\mathcal{\hat{G}}}, \mathbf{h}_{v_f}]) \cdot \mathrm{MLP}_K^v(\mathbf{h}_v)\right) \\
    P_e &= \underset{e\in \mathcal{E}_{\mathcal{W}}}{\mathrm{softmax}} \left(\mathrm{MLP}_Q^e ([\mathbf{h}_{\mathcal{\hat{G}}}, \mathbf{h}_{e_f}]) \cdot \mathrm{MLP}_K^e(\mathbf{h}_e)\right)
\end{align}
where $\mathcal{\hat{G}} = \mathcal{G}^{(t-1)}$. The type of FCS (atom or bond) is determined to be consistent with FCS in $\mathcal{G}^{(t-1)}$ to avoid mismatched connection. Up to now, the motif to be attached $\mathcal{W}^{(t-1)}$ is also determined and the new molecule  $\mathcal{G}^{(t)}$ is generated by attaching $\mathcal{W}^{(t-1)}$ to $\mathcal{G}^{(t-1)}$ on the FCSs of both. To realize the conformation of $\mathcal{G}^{(t)}$, we first represent the conformation of $\mathcal{G}^{(t-1)}$ as $C_{\mathcal{G}^{(t-1)}}$, and sample a conformation of $\mathcal{G}^{(t)}$ that denoted as $\hat{C}_{\mathcal{G}^{(t)}}$ with RDKit~\cite{Bento2020_RDKit}, then we use Kabsch algorithm~\cite{Kabsch1976_Kabsch} to calculate the translation $\mathbf{t}$ and the rotation $\mathbf{R}$ that align conformation of $\mathcal{W}^{(t-2)}$ in $\hat{C}_{\mathcal{G}^{(t)}}$ to the conformation of $\mathcal{W}^{(t-2)}$ in $C_{\mathcal{G}^{(t-1)}}$. Let $\mathbf{x}_l^i$ denote the position vector of atom $i$ in $\mathcal{W}^{(t-1)}$, its position after attachment $\mathbf{\hat{x}}_l^i$ is calculated as:
\begin{equation}
    \mathbf{\hat{x}}_l^i = \mathbf{R} \mathbf{x}_l^i + \mathbf{t}. 
\end{equation}
It should be noted that there may be additional freedom if the newly formed bond is rotatable, then the torsional angle should be predicted, which will be introduced in the next section.
     
\subsubsection{Torsional angle prediction}
\label{sec:torsional}
Torsional angle prediction is the last but important module that determines the conformation of the generated molecules, since bond lengths, bond angles and small rings are essentially rigid, such that the flexibility of molecules lies almost entirely in the torsional angles at rotatable bonds, and it is also hard to learn due to the flexibility. Previous approaches~\cite{Zhang2022_FLAG, Zhang2023_DrugGPS} predict the change of the torsional angle with a regression model, which tends to learn an average-like implausible output and may lead to unrealistic conformation. Inspired by the results achieved in~\cite{Jing2022_TorsionalDiff}, we propose a diffusion-based model to characterize the distribution of torsional angles. To be specific, for bond $b$, an $SE(3)$-invariant scalar representing torsional score $T_b$ is generated by a convolution of each atom with the bond center $z$:

\begin{equation}
    {T}_b = \mathrm{MLP}\left(\frac{1}{|\mathcal{N}_b|} \sum_{a\in\mathcal{N}_b} Y(\mathbf{r}_{za}) \otimes Y^2(\mathbf{r}_b) \otimes_{\gamma_{za}} \mathbf{h}_a \right), 
\end{equation}
where $\gamma_{za} = \Gamma(\mathrm{h}_{za}, \mathrm{h}_a, \mathrm{h}_{i}+\mathrm{h}_{j}), (v_i, v_j \in e_b)$, where $e_b$ is the edge of bond $b$.

\subsection{Training}
\label{training}
In the training stage, we use binary cross-entropy loss $\mathcal{L}_{fro}$ for the prediction of frontiers in pockets and cross-entropy loss $\mathcal{L}_{mot}$ for the first motif prediction, while binary cross entropy loss $\mathcal{L}_{CS}$ is used for connection site prediction, furthermore, $\mathcal{L}_{\mathbf{tr}}$, $\mathcal{L}_{\mathbf{rot}}$, $\mathcal{L}_{T}$ are the losses for the translation, rotation of the first motif and torsion of rotatable bond produced in generation. The total loss can be defined as follows:
\begin{equation}
    \mathcal{L} = \mathcal{L}_{fro} + \mathcal{L}_{mot} + \mathcal{L}_{CS} + \mathcal{L}_{\mathbf{tr}} + \mathcal{L}_{\mathbf{rot}} + \mathcal{L}_{T}. 
\end{equation}
We provide more implementation details in Appendix~\ref{appendix:implementation}.

\section{Experiment}

\subsection{Experiment Setup}
\noindent\textbf{Dataset}. In this paper, we train and evaluate our model with the CrossDock2020~\cite{Francoeur2020_CrossDock} dataset, which contains 22.5 million poses of ligands docked into multiple similar binding pockets across the Protein Data Bank~\cite{Berman2000_PDB}. In our experiments, the dataset is processed with the same procedure to~\cite{Guan2023_DecompDiff}.

\textbf{Baselines}. We compare our model with various state-of-the-art baselines: LiGAN~\cite{Ragoza2022_LiGAN} is a
 conditional variational autoencoder-based generation model. GraphBP~\cite{Liu2022_GraphBP}, AR~\cite{Luo2021_AR}, and Pocket2Mol~\cite{Peng2022_Pocket2Mol} are atom-wise autoregressive generation approaches. FLAG~\cite{Zhang2022_FLAG} generates molecules fragment-by-fragment in an autoregressive fashion. TargetDiff~\cite{Guan2023_TargetDiff} and DecompDiff~\cite{Guan2023_DecompDiff} are diffusion-based generation methods.

\textbf{Evaluation}. 
To compare with the existing state-of-the-art generation models more fairly and practically, we improve the evaluation framework by constraining the molecular weights of the generated molecules in the same range (detailed in Appendix~\ref{appendix:molwt}), which is different from previous evaluations since there is a strong correlation between Vina Score and molecular weight~\cite{Xu2022_Systematic}. Specifically, we evaluate the generated molecules from three perspectives: \textbf{(1) molecular structure validity}: we analyze the atom distance and bond angle respectively first, by calculating the \textit{Jensen-Shannon divergences (JSD)} between the generated molecules and the reference set. In addition, we also calculate the JSD between the generated molecules and the force-filed optimized ones, which does not rely on a specific reference set and achieves a more generalized and realistic estimation. Furthermore, to evaluate the whole structure comprehensively, we propose a new metric called conformer RMSD, which is specified in Section~\ref{sec:eval_structure}. \textbf{(2) pharmaceutical properties}: we choose the two most commonly used metrics which are also important reference indicators for pharmaceutical chemists in practical development: Synthetic Accessibility (SA) and Quantitative Estimation of Drug-likeness (QED), which follow the setup of~\citet{Guan2023_DecompDiff}. \textbf{(3) binding affinity}: we also evaluate the binding affinity of the generated molecules with AutoDock Vina~\cite{Eberhardt2021_VINA}. Following the setup of \citet{Guan2023_DecompDiff}, we report both the mean and the median value of four metrics: Vina Score, Vina Min, Vina Dock, and High Affinity. Additionally, we propose another two metrics, i.e., Vina Score$^{*}$ and Vina Min$^{*}$, which are specified in Section~\ref{sec:eval_affinity}.

\begin{figure}[t]
\vskip 0.2in
\begin{center}
\centerline{\includegraphics[width=\columnwidth]{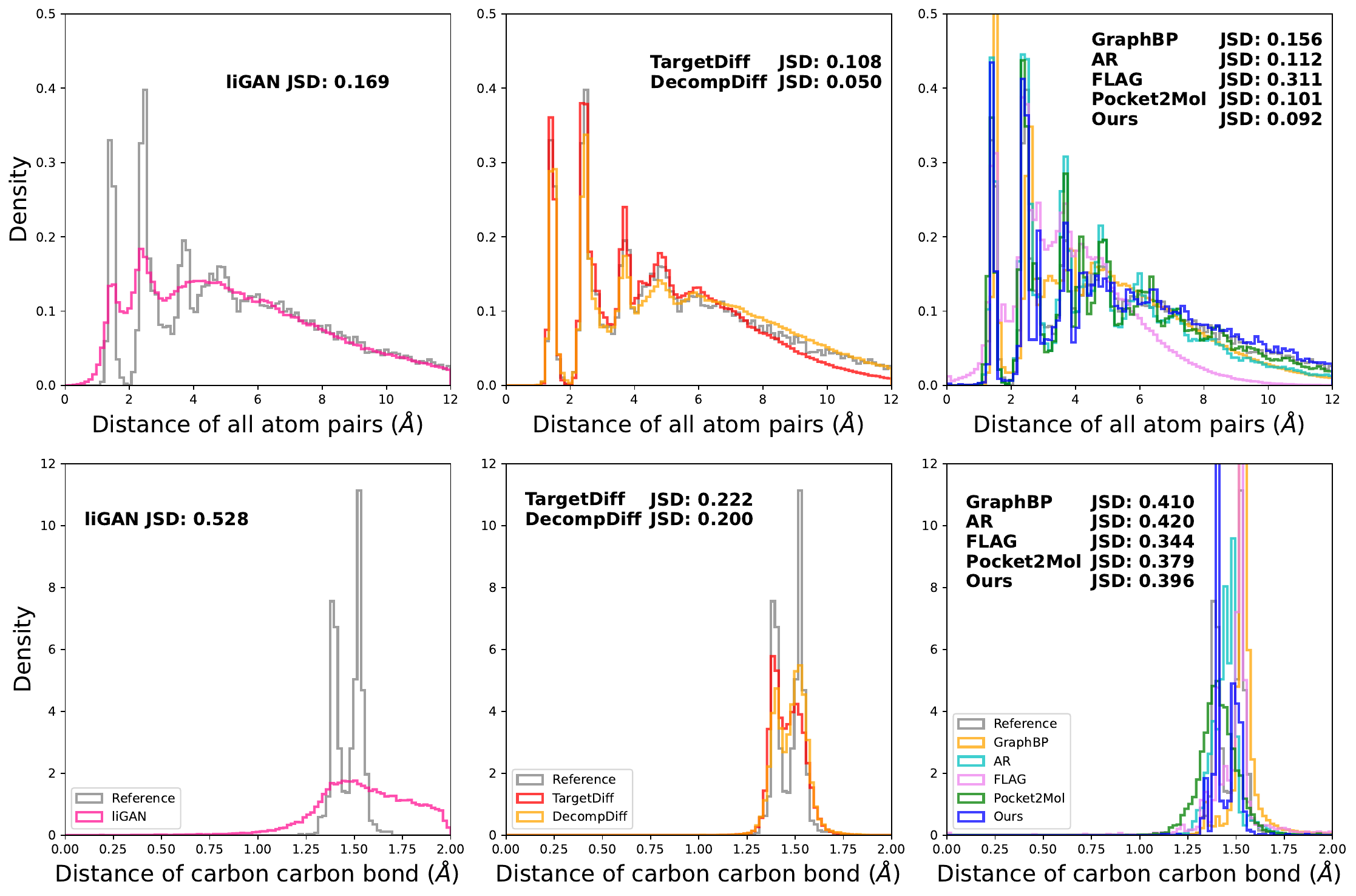}}
\caption{Comparing the distribution for distances of all-atom (top row) and carbon-carbon pairs (bottom row) for reference molecules (gray) and model generated molecules (color). JSD between two distributions is reported.}
\label{fig:distance}
\end{center}
\vskip -0.2in
\end{figure}

\subsection{Molecular Structure Validity Analysis}
\label{sec:eval_structure}
Firstly, we evaluate the structure validity by analyzing the distributions of all-atom distances and carbon-carbon pair distances and comparing them against the corresponding reference empirical distributions in Figure~\ref{fig:distance}. For overall atom distances, \mname~achieves the lowest JSD compared to other autoregressive-based approaches and competitive performance compared to diffusion-based approaches, which are similar to the results of carbon-carbon pair distances scenario. In addition, we compute the bond angle distributions of the generated molecules and compare them against the reference set (Table~\ref{table:jsd}, top rows), and similarly, \mname~achieves comparable performance as well.

\begin{figure}[ht]
\vskip 0.2in
\begin{center}
\centerline{\includegraphics[width=.8\columnwidth]{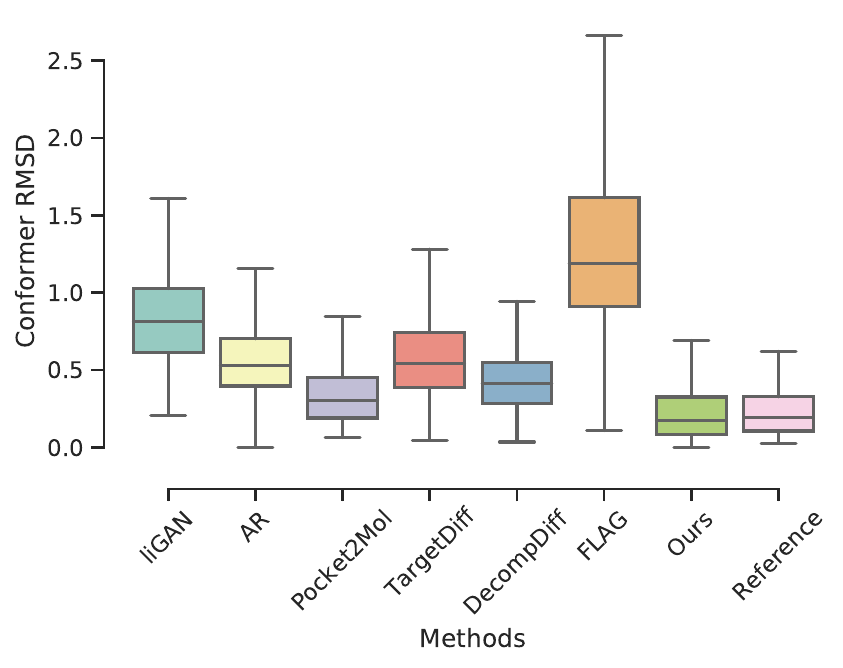}}
\caption{Conformer RMSD of the molecules sampled from different models. }
\label{fig:conformal-rmsd}
\end{center}
\vskip -0.2in
\end{figure}

Nevertheless, the capacity of JSD computed against the reference empirical distributions is limited to evaluate the structure validity exactly, since it prefers molecules that are structure-similar to the reference set, while not the ones that are dissimilar to the reference set but actually structure-valid. Therefore, we propose to compute JSD of the generated molecules against their force field optimized results rather than the reference set, as the outputs of the force field are generally considered to be approximate correct structures. The new metric is more generalized and alleviates the bias arose in the evaluation before. As shown in Table~\ref{table:jsd} (bottom rows), \mname~outperforms other baselines and achieves the best performance, and the results of JSD are close to the ones of the reference set, which means \mname~is capable of learning molecule structures with valid bond angles.

\begin{table}[tb]
\caption{JSD between bond angle distributions of the reference and the generated molecules (top), and of the generated and the force-field optimized molecules (bottom). The best two results are highlighted with \textbf{bold text} and \underline{underlined text}, respectively.}
\label{table:jsd}
\vskip 0.15in
\begin{center}
\resizebox{\columnwidth}{!}{%
\begin{tabular}{cccccccccc}
\toprule
Angle & Ref & \makecell{li\\GAN} & \makecell{Graph\\BP} & AR & \makecell{Pocket\\2Mol} & \makecell{Target\\Diff} & \makecell{Decomp\\Diff} & FLAG & Ours \\
\midrule
CCC & - & 0.60 & 0.38 & 0.33 & 0.34 & 0.33 & \bftab0.26 & 0.40 & \und{0.31} \\
CCO & - & 0.64 & 0.31 & 0.45 & 0.40 & 0.38 & \und{0.29} & 0.44 & \bftab0.27 \\
CNC & - & 0.62 & 0.45 & 0.38 & \bftab0.24 & 0.37 & \und{0.29} & 0.53 & 0.42 \\
NCC & - & 0.63 & 0.32 & 0.40 & 0.36 & 0.35 & \bftab0.25 & 0.44 & \und{0.32} \\
CC=O & - & 0.65 & 0.36 & 0.48 & 0.36 & 0.36 & \bftab0.25 & 0.45 & \und{0.30} \\
\midrule
CCC & 0.14 & 0.49 & \und{0.23} & 0.32 & 0.31 & 0.36 & 0.38 & 0.31 & \bftab0.22 \\
CCO & 0.20 & 0.62 & \und{0.31} & 0.47 & 0.43 & 0.44 & 0.43 & 0.40 & \bftab0.24 \\
CNC & 0.20 & 0.46 & \und{0.25} & 0.31 & 0.29 & 0.28 & 0.31 & 0.29 & \bftab0.23 \\
NCC & 0.20 & 0.52 & \bftab0.20 & 0.37 & 0.34 & 0.34 & 0.35 & 0.32 &\und{0.25} \\
CC=O & 0.30 & 0.64 & \und{0.24} & 0.56 & 0.47 & 0.44 & 0.34 & 0.38 & \bftab0.23 \\
\bottomrule
\end{tabular}
}
\end{center}
\vskip -0.1in
\end{table}

\begin{table*}[t]
\small
\caption{Results of binding affinities and pharmaceutical properties. Top 2 results are highlighted with \textbf{bold text} and \underline{underlined text}, respectively.}
\label{table:main}
\vskip 0.00in
\begin{center} 
\resizebox{\textwidth}{!}{%
\begin{tabular}{l|cc|cc|cc|cc|cc|cc|cc|cc}
\toprule
\multirow{2}{*}{Methods} & \multicolumn{2}{c}{Vina Score(↓)} & \multicolumn{2}{c}{Vina Score$^*$(↓)} & \multicolumn{2}{c}{Vina Min(↓)} & \multicolumn{2}{c}{Vina Min$^*$(↓)} & \multicolumn{2}{c}{Vina Dock(↓)} & \multicolumn{2}{c}{High Affinity(↑)} & \multicolumn{2}{c}{QED (↑)} & \multicolumn{2}{c}{SA (↑)} \\
 & Avg. & Med. & Avg. & Med. & Avg. & Med. & Avg. & Med. & Avg. & Med. & Avg. & Med. & Avg. & Med. & Avg. & Med. \\
 \midrule
Reference & -6.36 & -6.46 & -5.65 & -5.94 & -6.71 & -6.49 & -6.32 & -6.18 & -7.45 & -7.26 & -- & -- & 0.48 & 0.47 & 0.73 & 0.74 \\
\midrule
LiGAN & -- & -- & -- & -- & -- & -- & -- & -- & -8.49 & -8.39 & 0.64 & 0.69 & 0.35 & 0.30 & 0.54 & 0.52 \\
GraphBP & -- & -- & -- & -- & -- & -- & -- & -- & -2.49 & -3.96 & 0.10 & 0.03 & 0.49 & 0.50 & 0.49 & 0.49 \\
AR & -4.98 & \und{-6.40} & -3.37 & -4.23 & -6.51 & -6.76 & -5.33 & -5.53 & -7.67 & -7.40 & 0.58 & 0.69 & 0.47 & 0.46 & 0.56 & 0.55 \\
Pocket2Mol & \bftab-6.37 & \bftab-6.56 & \und{-4.72} & \und{-4.88} & \bftab{-7.39} &\bftab-7.54 & \und{-5.98} & \und{-6.26} & \und{-8.58} & \und{-8.63} & \und{0.68} & \bftab0.79 & \und{0.54} & \und{0.54} & \und{0.71} & \und{0.71} \\
FLAG & 51.03 & 42.13 & 50.08 & 41.90 & 9.42 & -2.23 & 8.63 & -2.12 & -5.49 & -6.04 & 0.26 & 0.10 & 0.35 & 0.31 & 0.49 & 0.48 \\
TargetDiff & \und{-5.83} & -6.36 & -2.64 & -3.79 & -6.87 & -6.89 & -4.50 & -4.84 & -7.85 & -7.94 & 0.60 & 0.60 & 0.50 & 0.50 & 0.59 & 0.58 \\
DecompDiff & -3.76 & -4.72 & -2.33 & -3.63 & -5.29 & -5.59 & -4.34 & -4.86 & -7.03 & -7.17 & 0.37 & 0.24 & 0.44 & 0.43 & 0.68 & 0.68 \\
\mname & -5.25 & -5.33 & \bftab-5.02 & \bftab-5.18 & \und{-6.91} & \und{-7.06} & \bftab-6.69 & \bftab-6.83 & \bftab-8.86 & \bftab-8.94 & \bftab0.73 & \und{0.77} & \bftab0.57 & \bftab0.58 & \bftab0.76 & \bftab0.77 \\
\bottomrule
\end{tabular}
}
\end{center}
\vskip -0.1in
\end{table*}

To further evaluate the structure validity comprehensively in addition to separate analysis of atom distances and bond angles, we design another new metric conformer RMSD inspired by the conformer matching~\cite{Jing2022_TorsionalDiff}: for a molecule $\mathcal{G}$ we optimize its conformation $C_{\mathcal{G}}$ by force-field to obtain $C_{\mathcal{G}}^{\mathcal{FF}}$, then we modify torsion angels of the $C_{\mathcal{G}}^{\mathcal{FF}}$ to match $C_{\mathcal{G}}$. The optimal match $(\hat{C}_{\mathcal{G}}^{\mathcal{FF}}, C_{\mathcal{G}})$ can be found by running a differential evolution optimization procedure over the torsion angles, and $\mathrm{RMSD} (\hat{C}_{\mathcal{G}}^{\mathcal{FF}}, C_{\mathcal{G}})$ is defined as conformer RMSD. As shown in Figure~\ref{fig:conformal-rmsd}, \mname~achieves the lowest conformer RMSD compared to all other baselines, which is also close to the result of the reference set. The result suggests that \mname~can generate molecules with more valid structures and conformations.

\begin{figure}[tb]
\vskip 0.2in
\begin{center}
\centerline{\includegraphics[width=.9\columnwidth]{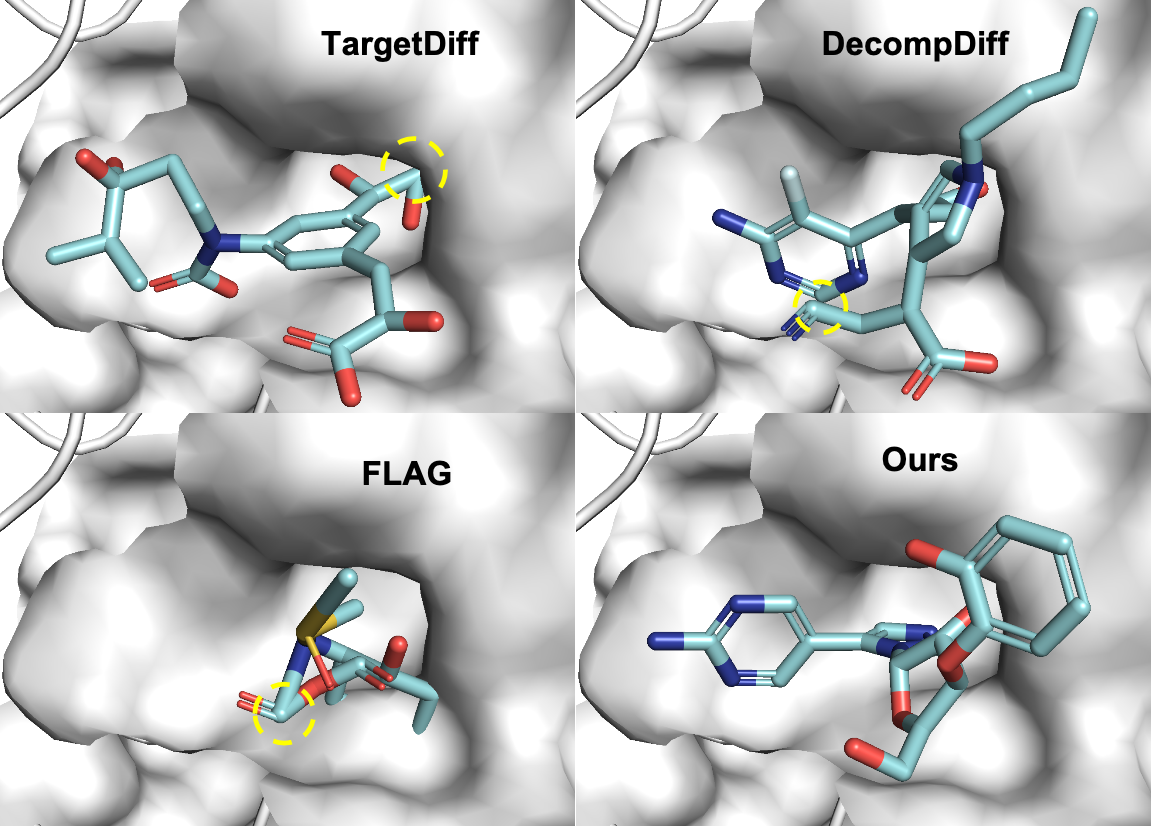}}
\caption{Visualization of chemically implausible local structures generated by TargetDiff, DecompDiff, FLAG. Incorrect bond angles are marked by yellow circles (PDBID is 2Z3H).}
\label{fig:case-study}
\end{center}
\vskip -0.2in
\end{figure}

\noindent\textbf{Case study.} To better understand the structure validity, a case study is conducted and reported in Figure~\ref{fig:case-study}. We can see that FLAG tends to generate unrealistic structures, and TargetDiff as well as DecompDiff generate approximate valid structures but incorrect local details such as invalid bond angles, while \mname~can generate rational molecules with more realistic structures and conformations, which can be attributed to the superiority of the conformal motif strategy.

\subsection{Binding Affinities and Pharmaceutical Properties}
\label{sec:eval_affinity}
In accordance with practice, we evaluate the binding affinity by computing Vina Score, Vina Min, Vina Dock and High Affinity first. It should be noted that our experiments are conducted under the constraint of molecular weights. In Table~\ref{table:main}, we can see that Pocket2Mol outperforms other models, while \mname~achieves competitive performance similar to Pocket2Mol and is better than other baselines.

However, the metrics Vina Score and Vina Min are not robust enough since they do not take the structure validity into account, which means molecules with unrealistic structures may still achieve decent Vina scores. To address this issue, we propose another two metrics, i.e., Vina Score$^{*}$ and Vina Min$^{*}$, which compute Vina scores for the molecules that are preprocessed with conformer matching~\cite{Jing2022_TorsionalDiff} rather than the ones generated by models. These two new metrics can evaluate the binding affinity more practically which ensure the approximate correctness of molecular local structures and conformations (bond lengths and bond angles). Table~\ref{table:main} shows that results of Vina Score$^{*}$ and Vina Min$^{*}$ are worse than Vina Score and Vin Min for almost all the models, which are reasonable since the metrics are more strict than before due to taking structure validity into account when docking, while we can see that \mname~achieves the best performance and the Vina score values also fall into a good range. It is unexpected that Pocket2Mol acquires the second-best results which are also very impressive. Furthermore, \mname~also obtains the highest QED and SA scores. All the results again suggest that \mname~is suitable for SBDD task and it can generate more drug-like molecules with good binding affinities.

\section{Conclusion}
In this paper, we propose \mname, a diffusion-based fragment-wise autoregressive generation approach, which can generate realistic molecules with valid structures and conformations based on the conformal motif. Moreover, we also improve the evaluation framework of SBDD, which can benchmark the generation models fairly and practically. In future work a fine-tuning module could be introduced to refine the intermediates during generation.

\section*{Impact Statement}

This paper presents work whose goal is to advance the field of Machine Learning. There are many potential societal consequences of our work, none of which we feel must be specifically highlighted here.

\bibliography{example_paper}
\bibliographystyle{icml2024}

\newpage
\appendix
\onecolumn

\section{Terminologies and notations}
\label{appendix:term}
We list terminologies and notations in Table~\ref{table:notation}
\begin{table}[h]
\caption{Terminologies and notations.}
\label{table:notation}
\vspace{0.2in}

\resizebox{\columnwidth}{!}{
\begin{tabular}{c|p{0.84\columnwidth}}
\toprule[1pt]
Notations & Explanations \\ 
\hline 
CCS & candidate connection sites  \\ 
RCCS & reduced candidate connection sites \\
FCS & focal connection site \\ 
$\mathcal{P} $ & protein pocket \\
$a_P^{i}$ & the atom type of the $i$-th atom in protein pocket \\ 
$\mathbf{r}_P^{i}$ & the coordinate of the $i$-th heavy atom in protein pocket \\
$\mathcal{G} $ & drug molecule (ligand) \\
$a_G^{i}$ & the atom type of the $i$-th atom in drug molecule \\ 
$\mathbf{r}_G^{i}$ & the coordinate of the $i$-th heavy atom in drug molecule \\ 
$\mathcal{W}$ & motif vocabulary \\ 
$\mathcal{W}_r$ & ring-like motif vocabulary \\ 
$\mathcal{W}_{c*}$ & chain-like motif vocabulary \\ 
$\mathcal{G}^{(t)}$ & drug molecule after generating $t$ motifs \\ 
$\phi$ & generation model \\ 
$\mathcal{G}_H$ & $\mathcal{G}_H = (\mathcal{V}_H, \mathcal{E}_H)$, heterogeneous geometric graphs \\ 
$\mathcal{V}_H$ & $\mathcal{V}_H = (\mathcal{V}_l, \mathcal{V}_p)$,  collection of all the heavy atoms of ligand fragment (l) and protein pocket (p) \\ 
$\mathcal{E}_H$ & edge set, $\mathcal{E}_H=(\mathcal{E}_{ll}, \mathcal{E}_{lp}, \mathcal{E}_{pl}, \mathcal{E}_{pp})$ \\ 
$\mathrm{MLP}$ & multiple layer perceptron \\ 
$\mathbf{h}_i$ & node $i$'s representation \\ 
$\mathrm{h}_i$ & node $i$'s scalar features \\ 
$\mathring{A}$ & Angstrom ($10^{-10}$ m) \\ 
$\oplus$ & vector addition \\ 
$\otimes_{\psi_{}}$ & spherical tensor product with weight $\psi_{}$ \\ 
$\mathrm{BN}$ & batch normalization \\ 
$\mathcal{N}_i^{(c)}$ & neighbors of node $i$ of type $c$ in radius graphs. \\ 
$c_a$ & atom type indicating whether the atom is in ligand or protein, $c_a \in \{l,p\}$. \\ 
$\mathcal{W}^{(t)}$ & the added motif at the $t$-th iteration \\ 
$\mathbf{R}$ & rotation matrix \\ 
$||\cdot||$ & $l_2$ norm of a vector \\ 
$T_b$ & torsion score of bond $b$ \\ 
\bottomrule[1pt]
\end{tabular}}

\end{table}

\section{Details in Conformal Motif Extraction}
\subsection{Definition of Freely Rotatable Bond}
\label{appendix:rotatable_bond}
In this paper, the freely rotatable bond is defined as follows: if cutting a bond creates two connected components of the molecule, and each connected component has at least one atom that is not in the direction of the severed bond, then the bond is considered to be freely rotatable. We only count single bonds as rotatable. Different from previous definitions~\cite{Jing2022_TorsionalDiff, Zhang2022_FLAG}, our definition guarantees that a freely rotatable bond is chemically rotatable and changes molecular conformation as it rotates.  

\subsection{Example of conformal motif extraction}
\label{appendix:motif_example}
\begin{figure}[h]
\vskip 0.2in
\begin{center}
\centerline{\includegraphics[width=.7\columnwidth]{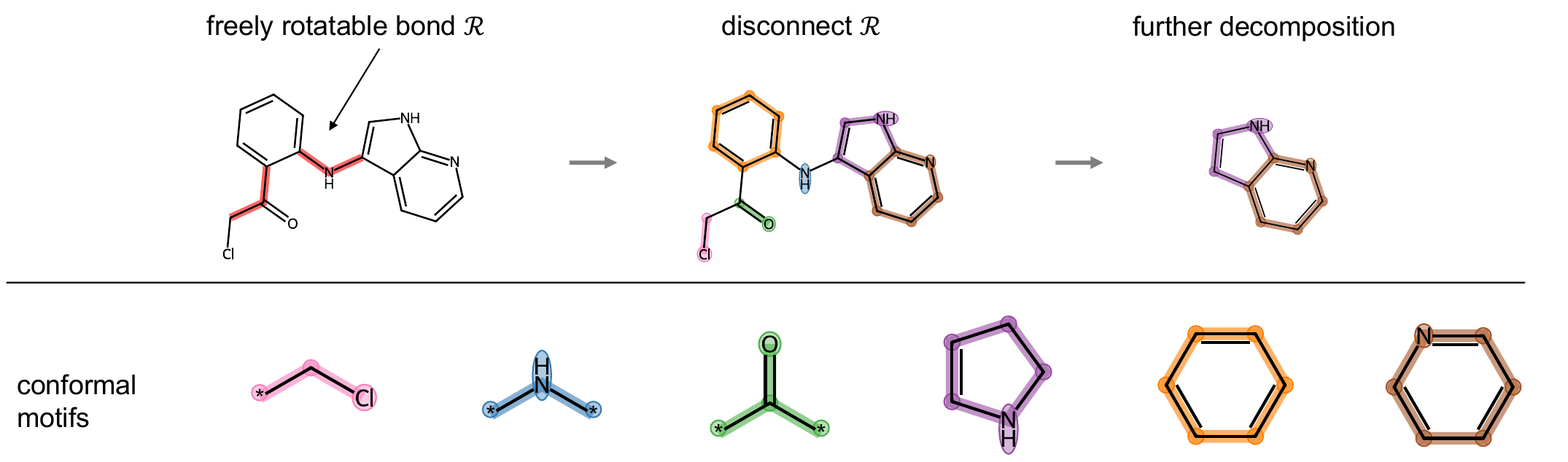}}
\caption{An example of conformal motif extraction, the symbol ‘*’ in motifs represents a dummy atom.}
\label{fig:motif_extract}
\end{center}
\vskip -0.2in
\end{figure}

In Figure~\ref{fig:motif_extract}, we provide an example of conformal motif extraction.

\section{Details of Molecular Weight Constraint}
\label{appendix:molwt}
By analysis we found that the molecular weights of molecules generated by different models always vary differently, autoregressive-based approaches tend to generate small molecules in an atom-by-atom (or motif-by-motif) fashion, while diffusion-based approaches generate molecules in a one-shot fashion, they determine the number of atoms before generation, thus allowing for more flexible control of the molecular weight. Taking the correlation between vina score and molecular weight into consideration, we constrain the molecular weights of the generated molecules in the same range to conduct a fair evaluation. Considering the statistical number of generated molecules and the similar molecular weight distribution between the molecules generated by TargetDiff and the reference set, for each testing protein pocket $\mathcal{P}_i$, we drop the top 20\% and the bottom 20\% of the molecules generated by TargetDiff according to the molecular weight and calculate the mean $\mu_i$ and standard deviation $\sigma_i$ of the remaining molecules. Then we define the valid molecular weight range $H_{i}$ as $[\mu_{i}-\sigma_{i}, \mu_{i}+\sigma_{i}]$. For protein pocket $\mathcal{P}_i$, only the molecules whose molecular weight fall in the range $H_{i}$ will be evaluated.

For each model, 100 molecules are generated and sampled in the range $H_{i}$, which are used to be evaluated in our experiments. Table~\ref{table:molwt} shows the molecular weights of molecules generated by various SBDD models under default settings and the molecular weight constraint settings.

\begin{table}[h]
\caption{Molecular weights for various models under default settings (MolWt1) and the molecular weight constraint settings (MolWt2).}
\label{table:molwt}
\vskip 0.1in
\begin{center}
\resizebox{\columnwidth}{!}{%
\begin{tabular}{lcccccccc}
\toprule
 & liGAN & GraphBP & AR & Pocket2Mol & FLAG & TargetDiff & DecompDiff & AutoDiff \\
 \midrule
MolWt1 & 294.87±25.20 & 344.54±171.75 & 250.48±58.84 & 242.75±51.99 & 287.33±81.07 & 347.34±85.03 & 581.92±42.30 & 254.62±60.98 \\
MolWt2 & 348.29±14.01 & 336.59±21.89 & 328.95±15.68 & 335.47±15.45 & 337.57±20.91 & 336.71±21.72 & 335.36±21.44 & 331.60±18.39\\
\bottomrule
\end{tabular}
}
\end{center}
\vskip -0.1in
\end{table}

\section{Implementation Details}
\label{appendix:implementation}
For node features of molecules, we use atom symbol, formal charge, number of explicit Hs, number of total Hs, and hybridization type. For node features of protein atoms, we use element types, the amino acids they belong to, and whether they are backbone or side-chain atoms. Edge features include the distances encoded with radial basis functions and bond type. The input scalar features of nodes and edges are concatenated with sinusoidal embeddings of diffusion time. 

In the training stage, we first construct motif trees of molecules, then we traverse motif trees in a breadth-first (BFS) order to get a traverse sequence $S$. We sample a mask ratio from the uniform distribution $\mathrm{U} [0, 1]$ and mask the corresponding number of the last $K$ motifs in $S$. Connection sites are determined during the masking procedure. 


\end{document}